\pdfoutput=1

\documentclass[11pt]{article}

\usepackage[preprint]{acl}

\usepackage{times}
\usepackage{latexsym}

\usepackage[T1]{fontenc}

\usepackage[utf8]{inputenc}

\usepackage{microtype}

\usepackage{inconsolata}

\usepackage{amsfonts,amsmath,amssymb}
\usepackage{graphicx}
\usepackage{multirow,multicol}
\usepackage{paralist}
\usepackage{booktabs}
\usepackage{hyperref}
\usepackage{cleveref}
\usepackage{xspace}
\usepackage[normalem]{ulem}
\usepackage{enumitem}
\usepackage{makecell}
\usepackage{listings}
\usepackage{xcolor}
\usepackage{stfloats}
\usepackage{rotating}

\definecolor{codegreen}{rgb}{0,0.6,0}
\definecolor{codegray}{rgb}{0.5,0.5,0.5}
\definecolor{codepurple}{rgb}{0.58,0,0.82}
\definecolor{backcolour}{rgb}{0.95,0.95,0.92}

\lstdefinestyle{prompt_style}{
    frame=single,
    basicstyle=\ttfamily\scriptsize,
    backgroundcolor=\color{white},
    stringstyle=\color{black},
    commentstyle=\color{darkgreen}\slshape,
    stringstyle=\color{darkred},
    numberstyle=\tiny\color{codegray},
    emphstyle=\color{pink}\underbar,
    breakindent=0pt,
    escapeinside={(*@}{@*)},
    breakatwhitespace=true,
    breaklines=true,
    captionpos=b,
    keepspaces=true,
    numbersep=5pt,
    showspaces=false,                
    showstringspaces=false,
    showtabs=false,
    tabsize=2
}

\crefname{equation}{Eq.}{Eq.}
\crefname{section}{Sec.}{Sec.}

\newcommand{\ie}{\emph{i.e.,}\xspace}

\newcommand{\eg}{\emph{e.g.,}\xspace}

\newcommand{\counterfact}{\textsc{CounterFact}\xspace}
\newcommand{\mquake}{\textsc{MQuAKE-CF}\xspace}
\newcommand{\wikiupdate}{\textsc{WikiUpdate}\xspace}
\newcommand{\del}[1]{\xspace\footnotesize{#1}}

\title{
    AKEW: Assessing Knowledge Editing in the Wild
}

\author{
  Xiaobao Wu$^1$ \quad Liangming Pan$^2$\thanks{Corresponding authors.} \quad William Yang Wang$^3$ \quad Anh Tuan Luu$^1$\footnotemark[1] \\
  $^1$Nanyang Technological University \qquad
  $^2$University of Arizona \\
  $^3$University of California, Santa Barbara \\
  \texttt{xiaobao002@e.ntu.edu.sg}
  \qquad
  \texttt{liangmingpan@arizona.edu}
  \\
  \texttt{william@cs.ucsb.edu}
  \qquad
  \texttt{anhtuan.luu@ntu.edu.sg}
}

\begin{document}
\maketitle

\begin{abstract}
    Knowledge editing injects knowledge updates into language models to keep them correct and up-to-date.
    However, its current evaluations deviate significantly from practice:
    their knowledge updates solely consist of structured facts derived from meticulously crafted datasets, instead of practical sources---unstructured texts like news articles, and they often overlook practical real-world knowledge updates.
    To address these issues, in this paper we propose AKEW (Assessing Knowledge Editing in the Wild), a new practical benchmark for knowledge editing.
    AKEW fully covers three editing settings of knowledge updates: structured facts, unstructured texts as facts, and extracted triplets.
    It further introduces new datasets featuring both counterfactual and real-world knowledge updates.
    Through extensive experiments, we demonstrate the considerable gap between state-of-the-art knowledge-editing methods and practical scenarios.
    Our analyses further highlight key insights to motivate future research for practical knowledge editing.%
    ~\footnote{Our code and data are available at \url{https://github.com/bobxwu/AKEW}.}
\end{abstract}

\section{Introduction}
    Recently Large Language Models (LLMs) have revolutionized the NLP field and derived various applications \cite{radford2019language,openai2023gpt4}.
    But a critical challenge arises that their knowledge could become incorrect or outdated over time \cite{elazar2021measuring,cao2021knowledgeable,dhingra2022time}.
    For this challenge, fine-tuning LLMs like continual learning \cite{jang2021towards,ke2022continual,wang2023orthogonal,padmanabhan2023propagating} is feasible, but often requires intensive computational cost and may cause forgetting issues \cite{dong2022calibrating,mitchell2022memory}.
    Due to this, knowledge editing has been proposed \cite{sinitsin2020editable,zhang2024comprehensive}.
    Compared to early continual learning,
    knowledge editing seeks to inject target knowledge into language models efficiently at low cost
    and assesses whether the edited models recall these new knowledge \cite{zhu2020modifying,decao2021editing}.
    A wide range of knowledge-editing methods have been proposed and demonstrated their effectiveness \cite{tan2024massive,hu2024wilke}.

    \begin{figure}
    \centering
    \includegraphics[width=\linewidth]{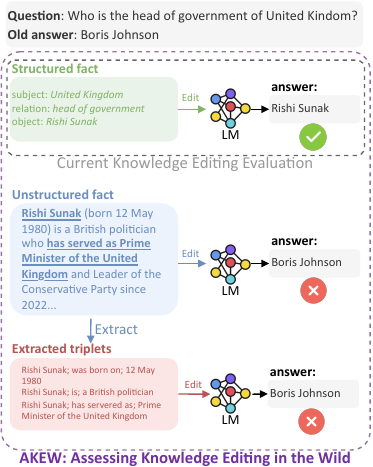}
    \caption{
        Illustration of current knowledge editing evaluation with only well-curated structured facts
        and our AKEW considering structured facts, unstructured facts, and extracted triplets.
        While knowledge-editing methods perform well on structured facts, they tend to fail on unstructured facts and extracted triplets.
    }
    \label{fig_motivation}
\end{figure}

    However, despite these achievements, we argue that current evaluations for knowledge editing deviate significantly from practice.
    First, their knowledge updates exclusively rely on well-curated structured facts---%
    a single isolated triplet $(s, r, o)$ with subject ($s$), relation ($r$), and object ($o$)---%
    sourced from meticulously crafted datasets \cite{wikidata2014,levy2017zero}.
    As such, they overlook the practical sources of knowledge updates---unstructured texts, such as scientific papers and news articles, where knowledge updates commonly emerge in practice \cite{tang2019learning}.
    Second, they often lack evaluations on practical real-world knowledge updates, while focusing on fabricated counterfacts \cite{meng2022locating,meng2022mass}.
    In consequence, current evaluations fail to consider the diversity and complexity challenge of practical knowledge editing.
    This inspires us to question: \textbf{\textit{How do current knowledge-editing methods perform in practice?}}

    To answer this question, we propose \textbf{AKEW} (\textbf{A}ssessing \textbf{K}nowledge \textbf{E}diting in the \textbf{W}ild),
    a new and comprehensive benchmark that evaluates knowledge-editing methods in practice.
    As shown in \Cref{fig_motivation}, AKEW covers three editing settings of knowledge updates for complete assessments:
    \begin{inparaenum}[(\bgroup\bfseries i\egroup)]
        \item
            \textbf{Structured fact}, including a single isolated triplet as in previous studies, like the one in \Cref{fig_motivation}.
        \item
            \textbf{Unstructured fact},
            an unstructured text containing the same knowledge in the structured fact,
            such as the biography of \textit{Rishi Sunak} in \Cref{fig_motivation}.
        \item
            \textbf{Extracted triplets} from the unstructured fact, like the triplets extracted from the biography in \Cref{fig_motivation}.
    \end{inparaenum}
    Based on the above settings, AKEW introduces three new datasets for evaluation:
    Two adapted from previous work include counterfactual knowledge updates with LLM-generated unstructured facts,
    and the new {\textbf{\wikiupdate}} contains real-world knowledge updates with unstructured facts retrieved from Wikipedia.

    We evaluate various state-of-the-art knowledge-editing methods.
    As exemplified in \Cref{fig_motivation}, we observe that they commonly excel on structured facts, but fail significantly on unstructured facts.
    We also find that extracted triplets are helpful to some knowledge-editing methods but they still fall short compared to structured facts.
    As a result, these experimental results disclose the considerable gap between existing knowledge-editing methods and the practical scenarios,
    thus emphasizing the urgency for more research into practical knowledge editing.
    The contributions of this paper can be concluded as follows:
    \begin{itemize}[leftmargin=*, parsep=0pt]
        \item
            We propose AKEW (Assessing Knowledge Editing in the Wild),
            a new comprehensive benchmark comprising three editing settings that fully evaluate knowledge editing in practice.
        \item
            We introduce three new datasets tailored for this benchmark, featuring both counterfactual and real-world knowledge updates.
        \item
            We conduct extensive experiments involving state-of-the-art knowledge-editing methods and language models,
            and reveal their general limitations in practical scenarios, which appeals for further research into practical knowledge editing.
    \end{itemize}

\section{Related Work}
    \paragraph{Knowledge-Editing Methods}
        Early knowledge-editing methods directly fine-tune model parameters with knowledge updates.
        For instance,
        \citet{zhu2020modifying} put a norm constraint on language model parameters during fine-tuning,
        preventing the model from forgetting original knowledge.
        \citet{hu2021lora} propose an efficient fine-tuning approach, LoRA, which trains low-rank matrices and freezes original model parameters.

        Others follow the Locate-Then-Edit principle based on the knowledge neuron view of Feed-Forward Networks
        \cite{geva2021transformer,geva2022transformer,gupta2024rebuilding}.
        They locate the parameters related to a knowledge edit and then modify them to inject new knowledge \cite{dai2022knowledge}.
        ROME \cite{meng2022locating} locates the parameters related to a knowledge edit
        and then modify them to inject new knowledge;
        MEMIT extends it by enabling mass editing \cite{meng2022mass}.
        Recently, \citet{li2023pmet} propose to refine the updating of FFN for more precise editing.

        Alternatively, other knowledge-editing approaches choose to preserve model parameters.
        \citet{decao2021editing} use hyper-networks to model the parameter updates of language models \cite{decao2021editing}.
        MEND \cite{mitchell2022fast} transforms the gradients computed by fine-tuning into a low-rank decomposition.
        Some studies follow memory-based strategies to avoid fine-tuning \cite{hartvigsen2022aging,mitchell2022memory}.
        SERAC \cite{mitchell2022memory} stores editing facts in memory and builds a counterfactual model to handle the input queries that fall within the editing scope.
        Following this line,
        IKE \cite{zheng2023ike} adopts in-context learning via editing exemplars and retrieves related facts in the memory.
        MeLLo \cite{zhong2023mquake} solves multi-hop editing by breaking down a multi-hop question into subquestions and retrieving related facts.
        As such, these two convert knowledge editing into a retrieval-augmented generation (RAG) task.

    \paragraph{Knowledge-Editing Evaluations}
        Current evaluations of knowledge editing mainly use well-curated structured facts as edits, each of which includes a single isolated triplet \cite{cohen2023evaluating,wei2024mlake}.
        These triplets originate from meticulously crafted datasets rather than practical sources.
        Recently \citet{wu2023evakellm} propose to edit with raw documents.
        However, they only include fine-tuning methods and lack the latest knowledge-editing methods;
        Besides, they merely consider counterfacts as edits, which still deviates from the practice.
        Different from these studies, our AKEW (Assessing Knowledge Editing in the Wild)
        covers three editing settings of knowledge updates: structured facts, unstructured facts, and extracted triplets.
        We experiment with state-of-the-art knowledge-editing methods (modifying or preserving parameters) on both counterfactual and real-world knowledge updates.
        These differences make our AKEW a more practical benchmark for knowledge editing.

\section{Assessing Knowledge Editing in the Wild}
    In this section,
    we analyze the problems of previous knowledge-editing evaluations
    and propose our new benchmark \textbf{AKEW} (\textbf{A}ssessing \textbf{K}nowledge \textbf{E}diting in the \textbf{W}ild).

    \subsection{Problems of Knowledge-Editing Evaluations}
        Knowledge editing seeks to inject knowledge updates into language models to keep them correct and up-to-date
        at low cost without expensive retraining \cite{yao2023editing,wang2023knowledge}.
        The target knowledge updates are denoted as edits.
        Current evaluations for knowledge editing merely use well-curated structured facts as edits---a single isolated triplet $(s, r, o)$ with subject $s$, relation $r$, and object $o$, for instance (\emph{United Kingdom}, \emph{head of government}, \emph{Rishi Sunak}) in \Cref{fig_motivation}.
        After editing, they verify whether edited language models can remember these facts by querying them through either a cloze-style statement like \emph{The head of government of United Kingdom is \_\_} \cite{meng2022locating,meng2022mass}, or a question like \emph{Who is the head of government of United Kingdom?} \cite{zheng2023ike,zhong2023mquake}.

\begin{figure}
    \centering
    \includegraphics[width=0.9\linewidth]{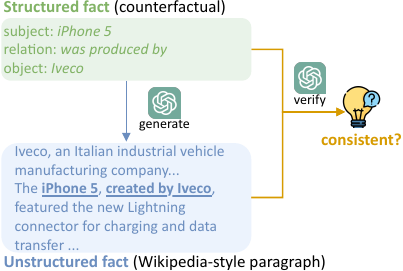}
    \caption{
        An example of generating Wikipedia-style paragraphs as unstructured facts for editing.
    }
    \label{fig_generation_unstruct}
\end{figure}

        Unfortunately, current evaluations deviate considerably from practice.
        In detail, they typically derive structured facts as knowledge updates from meticulously crafted datasets \cite{wikidata2014,levy2017zero}.
        As a result, they overlook the evaluations on the practical sources of knowledge updates, \ie unstructured texts like news articles, academic publications, and Wikipedia pages \cite{tang2019learning}.
        Knowledge updates frequently occur in them, such as presidential elections or corporate mergers and acquisitions.
        Besides, their edits mainly comprise outdated updates or fabricated counterfacts \cite{meng2022locating},
        \eg \textit{iPhone 5 was produced by Iveco} in \Cref{fig_generation_unstruct},
        so they lack practical real-world knowledge updates.
        Due to these issues, current evaluations ignore the diverse and complex nature of knowledge editing in practical scenarios,
        and it remains uncertain how current knowledge-editing methods perform in the wild.

\begin{figure*}[!t]
    \centering
    \includegraphics[width=\linewidth]{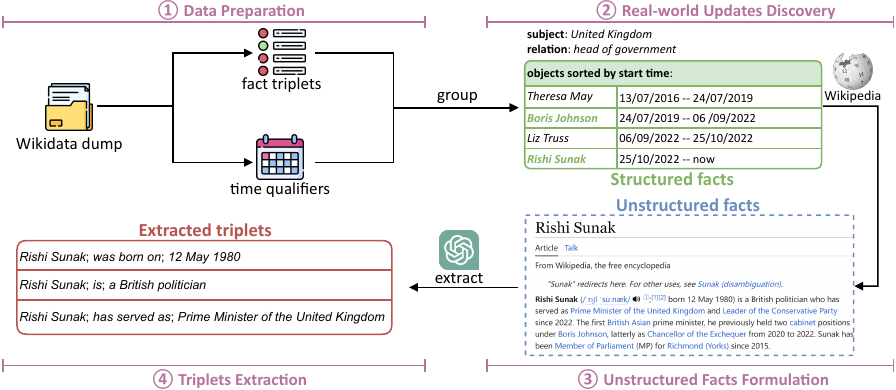}
    \caption{
        Construction process of \wikiupdate, including 4 steps:
        (1) Data Preparation;
        (2) Real-world Updates Discovery;
        (3) Unstructured Facts Formulation;
        (4) Triplets Extraction.
    }
    \label{fig_dataset_construction}
\end{figure*}

    \subsection{AKEW} \label{sec_AKEW}
        Motivated by the above analysis,
        we propose \textbf{AKEW},
        a more practical and comprehensive benchmark for knowledge editing.
        To evaluate knowledge editing in practice,
        AKEW considers the following three editing settings:
        \begin{itemize}[leftmargin=*,itemsep=0pt]
            \item
                \textbf{Structured Fact}.
                Following previous studies, each structured fact comprises a single isolated triplet for editing, which stems from existing datasets or knowledge databases.
            \item
                \textbf{Unstructured Fact}.
                We further use unstructured texts as facts for editing, denoted as unstructured facts.
                For fair comparisons, each unstructured fact covers the same knowledge update in its corresponding structured counterpart.
                Compared to structured ones, unstructured facts tend to exhibit higher complexity in the natural language format as they commonly contain more knowledge.
                For example, the biography of \textit{Rishi Sunak} in \Cref{fig_motivation} includes various details like his nationality, political position, and birthdate.
                We expect that an effective knowledge-editing method can edit such unstructured facts into a language model.
            \item
                \textbf{Extracted Triplets}.
                We finally use automatic methods to extract triplets from the unstructured fact,
                \eg extract the triplets from the biography of \textit{Rishi Sunak} as in \Cref{fig_motivation}.
                This is to investigate if extracted triplets can facilitate knowledge-editing methods to handle unstructured facts.
        \end{itemize}
        In short, these three settings provide comprehensive assessments of knowledge-editing methods' capabilities in practice,
        addressing the confines of previous well-curated edits alone.

\section{Dataset Construction for AKEW}
    In this section we construct three new datasets for the AKEW benchmark, two containing counterfactual updates and one featuring real-world updates.

    \subsection{Counterfactual Update Datasets}

        \paragraph{Data Preparation}
            We adopt two widely-used datasets using counterfactual updates as edits: \textbf{\counterfact} \cite{meng2022locating} and \textbf{\mquake} \cite{zhong2023mquake}.
            \counterfact manufactures counterfactual updates by replacing objects in triplets with similar but fake ones.
            \Cref{fig_generation_unstruct} illustrates the object of \emph{iPhone 5 was produced by \_\_} is updated from \emph{Apple} to \emph{Iveco}.
            Similarly, \mquake contains counterfactual updates but evaluates multi-hop editing.
            It measures whether edited language models can answer multi-hop questions as entailed consequences of edits.
            For instance, if we edit the UK Prime Minister, the spouse of the UK Prime Minister should change accordingly.

\begin{table}[!t]
    \centering
    \renewcommand{\arraystretch}{1.15}
    \resizebox{\linewidth}{!}{
    \begin{tabular}{ll}
    \toprule
    \textbf{Attributes} & \textbf{Examples} \\
    \midrule
    Subject & \emph{United Kingdom} \\
    \midrule
    Relation & \emph{head of government} \\
    \midrule
    Question & \makecell[l]{Who is the head of \\ \;\; government of United Kingdom?} \\
    \midrule
    Old object & \emph{Boris Johnson} \\
    $\quad \vdash$ start time & $\quad$ 2019-07-24 \\
    $\quad \vdash$ end time & $\quad$ 2022-09-06 \\
    \midrule
    Object & \emph{Rishi Sunak} \\
    $\quad \vdash$ start time & $\quad$ 2022-10-25 \\
    $\quad \vdash$ end time & $\quad$ Null \\
    \midrule
    Unstructured fact & \makecell[l]{Rishi Sunak is a British politician\\ who has served as Prime Minister \\ of the United Kingdom since 2022...} \\
    \midrule
    \multirow{3}[2]{*}{Extracted triplets} & \emph{Rishi Sunak; was born on; 12 May 1980} \\
          & \emph{Rishi Sunak; is; a British politician} \\
          & \emph{\makecell[l]{Rishi Sunak; has served as; Prime \\ \;\;  Minister of the United Kingdom}} \\
    \bottomrule
    \end{tabular}%
    }
    \caption{
        Data structure of \wikiupdate dataset.
    }
    \label{tab_wikiupdate_structure}%
\end{table}%

        \paragraph{Unstructured Facts Generation}
            Since both datasets exclusively consist of structured facts as edits, we need to construct their corresponding unstructured facts.
            As depicted in \Cref{fig_generation_unstruct}, we prompt ChatGPT \cite{ouyang2022training,openai2022chatgpt}
            to generate a Wikipedia-style paragraph describing a given structured fact while ignoring its information accuracy.
            This is due to several challenges stemming from the counterfactual nature of these datasets.
            We cannot retrieve evidence from the real world to support the counterfacts in them, \eg no evidence exists to support the counterfacts about iPhone 5 in \Cref{fig_generation_unstruct}.
            In addition, relying on human annotators to fabricate evidence for counterfacts introduces subjective biases and potential inaccuracies, since it is a counter-intuitive process and demands intricate and nuanced background knowledge.
            Leveraging ChatGPT can overcome these challenges due to its ability to generate coherent and contextually relevant texts
            and its inherent abundant knowledge learned from various domains.
            Note that we also construct real-world updates as unstructured facts later in \Cref{sec_wikiupdate}.

            We proceed to verify the consistency of the generated paragraphs.
            This is a necessary step due to the possibility that ChatGPT may decline to describe seriously incorrect counterfacts, such as \textit{United Kingdom is located in Asia}.
            As illustrated in \Cref{fig_generation_unstruct}, we prompt ChatGPT to verify whether the generated paragraph aligns with its original structured counterpart, and then manually check and filter out those unaligned ones.
            Then we consider filtered paragraphs as unstructured facts.
            See the prompts used for generation and verification in \Cref{app_prompts}.

    \subsection{Real-World Update Dataset} \label{sec_wikiupdate}
        Apart from the above counterfactual updates,
        we further build \textbf{\wikiupdate}, a novel dataset featuring knowledge updates in the real world for more practical evaluations.
        Previously \citet{zhong2023mquake} present an available real-world update dataset,
        but it solely comprises structured facts and is limited by its small scale and numerous repetitive samples.
        \Cref{fig_dataset_construction} illustrates the construction process, and \Cref{tab_wikiupdate_structure} details the data structure of \wikiupdate.

        \paragraph{Data Preparation}
            We use the dump of Wikidata \cite{wikidata2014} released in 09/2023 as our data source,
            which contains a vast range of real-world fact triplets, spanning millions of entities.
            To identify editing-worthy facts, we sample triplets according to their relation types.
            We retain the relations associated with physical entities and exclude virtual entities like ISBN or movie IDs
            (See \Cref{app_dataset_construction} for details).
            We also retrieve two qualifiers for each triplet from Wikidata---\textit{start time} and \textit{end time},
            indicating when the fact in the triplet starts and ends, such as the term of the UK Prime Minister in \Cref{fig_dataset_construction}.

        \paragraph{Real-world Updates Discovery}
            We subsequently discover real-world knowledge updates.
            Following \citet{zhong2023mquake}, we find the updates starting from 01/04/2021 (the update timestamp).
            Note that our discovery algorithm can alter this timestamp to discover updates for different editing purposes.
            The updates are indicated by the changes of objects \cite{decao2021editing};
            hence we group these triplets by subject and relation,
            and within each group, we identify the object with the latest start time and the object whose start time immediately precedes the update timestamp.
            If these two objects are different, this signifies that the object changes over time;
            thereby we regard this as a real-world update and the triplet with the latest object as a structured fact for editing.
            For instance, \Cref{fig_dataset_construction} step 2 illustrates the UK Prime Minister is \emph{Rishi Sunak} now and was \emph{Boris Johnson} before 01/04/2021.
            As such, we obtain the structured facts of real-world updates.

        \paragraph{Unstructured Facts Formulation}
            We retrieve Wikipedia to formulate unstructured facts.
            Given the triplet of an obtained structured fact,
            we retrieve the corresponding Wikipedia pages of its subject and object, and then extract their summaries (usually the first paragraphs) as unstructured facts.
            For instance, \Cref{fig_dataset_construction} step 3 shows the summary in \emph{Rishi Sunak}'s Wikipedia page.
            These summaries mostly describe the updates about the triplet, but they may not.
            Owing to this, we prompt ChatGPT to verify the alignment between the unstructured facts and their structured counterparts and manually check and filter out unaligned ones, similar to the process in \Cref{fig_generation_unstruct}.

\begin{table*}[!t]
    \centering
    \setlength{\tabcolsep}{2mm}
    \renewcommand{\arraystretch}{1.15}
    \resizebox{\linewidth}{!}{
        \begin{tabular}{cl|rrr|rrr|rrr}
        \toprule
        \multicolumn{1}{c}{\multirow{2}[4]{*}{\textbf{\makecell{Language \\ Model}}}} & \multirow{2}[4]{*}{\textbf{\makecell{Knowledge- \\ Editing Method}}} & \multicolumn{3}{c|}{\textbf{\counterfact}} & \multicolumn{3}{c|}{\textbf{\mquake}} & \multicolumn{3}{c}{\textbf{\wikiupdate}} \\
        \cmidrule{3-11}      &       & \textbf{Struct} & \multicolumn{1}{r}{\textbf{Unstruct}} & \multicolumn{1}{r|}{\textbf{Extract}} & \textbf{Struct} & \multicolumn{1}{r}{\textbf{Unstruct}} & \multicolumn{1}{r|}{\textbf{Extract}} & \textbf{Struct} & \multicolumn{1}{r}{\textbf{Unstruct}} & \multicolumn{1}{r}{\textbf{Extract}} \\
        \midrule
        \multirow{6}[6]{*}{\begin{sideways}GPT2-XL\end{sideways}} & \texttt{FT} & 97.33  & 0.07\del{↓100\%} & 11.49\del{↓88\%} & 38.30  & 0.23\del{↓99\%} & 4.13\del{↓89\%} & 5.16  & 0.09\del{↓98\%} & 0.28\del{↓95\%} \\
              & \texttt{LoRA} & 91.59  & 19.28\del{↓79\%} & 23.39\del{↓74\%} & 66.74  & 25.46\del{↓62\%} & 25.69\del{↓62\%} & 67.67  & 5.44\del{↓92\%} & 0.07\del{↓100\%} \\
        \cmidrule{2-11}      & \texttt{ROME} & 99.80  & \multicolumn{1}{c}{---} & 13.95\del{↓86\%} & 76.61  & \multicolumn{1}{c}{---} & 11.47\del{↓85\%} & 93.53  & \multicolumn{1}{c}{---} & 4.78\del{↓95\%} \\
              & \texttt{MEMIT} & 91.69  & \multicolumn{1}{c}{---} & 10.46\del{↓89\%} & 64.68  & \multicolumn{1}{c}{---} & 7.57\del{↓88\%} & 42.64  & \multicolumn{1}{c}{---} & 0.47\del{↓99\%} \\
        \cmidrule{2-11}      & \texttt{IKE (single)} & 79.18  & 72.72\del{↓8\%} & 46.97\del{↓41\%} & 82.80  & 63.53\del{↓23\%} & 46.33\del{↓44\%} & 97.38  & 56.23\del{↓42\%} & 28.77\del{↓70\%} \\
              & \texttt{IKE (all)} & 79.08  & 72.10\del{↓9\%} & 46.87\del{↓41\%} & 83.98  & 59.05\del{↓30\%} & 43.92\del{↓48\%} & 96.72  & 46.11\del{↓52\%} & 25.68\del{↓73\%} \\
        \midrule
        \midrule
        \multirow{6}[6]{*}{\begin{sideways}GPT-J (6B)\end{sideways}} & \texttt{FT} & 98.67  & 2.67\del{↓97\%} & 21.03\del{↓79\%} & 33.95  & 0.23\del{↓99\%} & 6.65\del{↓80\%} & 5.06  & 0.56\del{↓89\%} & 0.09\del{↓98\%} \\
              & \texttt{LoRA} & 85.95  & 17.85\del{↓79\%} & 21.23\del{↓75\%} & 76.15  & 19.73\del{↓74\%} & 22.25\del{↓71\%} & 91.47  & 5.16\del{↓94\%} & 4.22\del{↓95\%} \\
        \cmidrule{2-11}      & \texttt{ROME} & 99.59  & \multicolumn{1}{c}{---} & 16.72\del{↓83\%} & 80.28  & \multicolumn{1}{c}{---} & 16.28\del{↓80\%} & 99.63  & \multicolumn{1}{c}{---} & 2.33\del{↓98\%} \\
              & \texttt{MEMIT} & 99.49  & \multicolumn{1}{c}{---} & 16.31\del{↓84\%} & 81.19  & \multicolumn{1}{c}{---} & 11.93\del{↓85\%} & 99.81  & \multicolumn{1}{c}{---} & 2.25\del{↓98\%} \\
        \cmidrule{2-11}      & \texttt{IKE (single)} & 89.64  & 74.15\del{↓17\%} & 47.08\del{↓47\%} & 88.76  & 72.94\del{↓18\%} & 50.00\del{↓44\%} & 99.06  & 76.10\del{↓23\%} & 29.43\del{↓70\%} \\
              & \texttt{IKE (all)} & 89.54  & 73.54\del{↓18\%} & 46.97\del{↓48\%} & 86.94  & 68.55\del{↓21\%} & 46.29\del{↓47\%} & 98.41  & 61.95\del{↓37\%} & 25.87\del{↓74\%} \\
        \midrule
        \midrule
        \multirow{6}[6]{*}{\begin{sideways}Mistral (7B)\end{sideways}} & \texttt{FT} & 39.28  & 1.13\del{↓97\%} & 3.18\del{↓92\%} & 22.25  & 0.69\del{↓97\%} & 3.21\del{↓86\%} & 10.31  & 0.08\del{↓99\%} & 1.22\del{↓88\%} \\
              & \texttt{LoRA} & 90.87  & 5.85\del{↓94\%} & 9.74\del{↓89\%} & 52.29  & 6.65\del{↓87\%} & 3.21\del{↓94\%} & 25.86  & 0.04\del{↓100\%} & 0.09\del{↓100\%} \\
        \cmidrule{2-11}      & \texttt{ROME} & 76.00  & \multicolumn{1}{c}{---} & 6.46\del{↓91\%} & 60.55  & \multicolumn{1}{c}{---} & 3.67\del{↓94\%} & 23.99  & \multicolumn{1}{c}{---} & 0.44\del{↓98\%} \\
              & \texttt{MEMIT} & 80.21  & \multicolumn{1}{c}{---} & 10.36\del{↓87\%} & 63.76  & \multicolumn{1}{c}{---} & 3.37\del{↓95\%} & 55.95  & \multicolumn{1}{c}{---} & 0.37\del{↓99\%} \\
        \cmidrule{2-11}      & \texttt{IKE (single)} & 95.18  & 75.25\del{↓21\%} & 44.62\del{↓53\%} & 98.17  & 73.85\del{↓25\%} & 48.85\del{↓50\%} & 100.00  & 92.32\del{↓8\%} & 26.99\del{↓73\%} \\
              & \texttt{IKE (all)} & 95.08  & 74.67\del{↓21\%} & 44.51\del{↓53\%} & 97.92  & 70.33\del{↓28\%} & 45.40\del{↓54\%} & 99.34  & 73.20\del{↓26\%} & 23.34\del{↓77\%} \\
        \bottomrule
        \end{tabular}%
    }
    \caption{
        Editing accuracy of knowledge-editing methods on the counterfactual updates (\counterfact, \mquake), and real-world updates (\wikiupdate) under three settings:
        \textbf{Struct} (structured facts),
        \textbf{Unstruct} (unstructured facts),
        and \textbf{Extract} (extracted triplets).
        Percentages refer to the differences compared to the Struct setting.
    }
    \label{tab_editing}
\end{table*}%

    \subsection{Triplets Extraction} \label{sec_extract_triplets}
        We further extract triplets from the unstructured facts in the above three datasets for editing.
        This can investigate whether extracted triplets can assist knowledge-editing methods to handle unstructured facts.
        In detail, we employ ChatGPT to automatically extract all triplets from the unstructured facts within each dataset.
        Compared to the well-curated structured fact with a single triplet, this extraction process yields multiple triplets for each edit.
        For instance, \Cref{fig_dataset_construction} step 4 illustrates that the extracted triplets not only involve the Prime Minister but also other basic details regarding \emph{Rishi Sunak}.
        We opt not to employ filtering during this process, because we prefer more robust editing methods capable of editing all related triplets at once, instead of only supporting a single isolated triplet at each time.
        Additionally, filtering may inadvertently remove valuable knowledge updates, leaving LLMs inadequately edited.
        As such, this aligns more closely with the practical scenarios.
        See the prompt used for triplet extraction in \Cref{app_prompts}.

    \subsection{Dataset Summary} \label{sec_dataset_summary}
        We summarize the statistics of all three datasets in \Cref{tab_dataset_statistics}.
        Note that \wikiupdate includes longer unstructured facts compared to the other two, as well as more extracted triplets per edit.
        See more dataset details in \Cref{app_dataset_construction}.

\section{Experiment} \label{sec_experiment}
    In this section, we conduct experiments with the above datasets
    to evaluate current knowledge-editing methods on the AKEW benchmark.

    \subsection{Experiment Setup} \label{sec_experiment_setup}
        \paragraph{Base Language Models}
            Following previous studies \cite{zheng2023ike,zhong2023mquake},
            we use \textbf{GPT2-XL} \cite{radford2019language} and \textbf{GPT-J (6B)} \cite{gpt-j2021} as our base language models to be edited for experiments.
            We also adopt the more recent \textbf{Mistral (7B)} (released in Oct 2023), which outperforms Llama 2 (13B) as they report \cite{jiang2023mistral}.
            In \Cref{app_results_vicuna} we additionally experiment with Vicuna \cite{vicuna2023} as the base language model.

        \paragraph{Knowledge-Editing Methods}
            We first consider methods for \textbf{Continual Knowledge Learning} \cite{jang2021towards},
            which follow various fine-tuning strategies:
            \begin{inparaenum}[(i)]
                \item \texttt{\textbf{FT}} \cite{zhu2020modifying} fine-tunes under a norm constraint on model parameters to prevent from forgetting original knowledge.
                \item \texttt{\textbf{LoRA}} \cite{hu2021lora} trains low-rank matrices as alternatives for efficient fine-tuning.
            \end{inparaenum}

            Then we adopt the following knowledge-editing methods, categorized into two types:
            \begin{inparaenum}[(1)]
                \item
                    \textbf{Locate-Then-Edit}.
                    These locate the parameters related to knowledge in language models and edit them accordingly.
                    We use two representative methods:
                    \begin{inparaenum}[(i)]
                        \item \texttt{\textbf{ROME}} \cite{meng2022locating} locates the related feedforward network (FFN) in a language model and edits it to insert new knowledge.
                        \item \texttt{\textbf{MEMIT}} \cite{meng2022mass} extends \texttt{ROME} by updating FFN in a range of layers and enabling mass editing with a large set of knowledge.
                    \end{inparaenum}
                \item
                    \textbf{In-Context Learning}.
                    These preserve model parameters and retrieve new facts in memory for in-context learning.
                    We employ the latest methods:
                    \begin{inparaenum}[(i)]
                        \item \texttt{\textbf{IKE}} \cite{zheng2023ike} uses demonstration exemplars and retrieves stored facts in memory for in-context learning.
                        \item \texttt{\textbf{MeLLo}} \cite{zhong2023mquake} focuses on multi-hop knowledge editing.
                        It decomposes a multi-hop question into subquestions \cite{zhou2023leasttomost} and adjusts answers
                        by retrieving new facts.
                    \end{inparaenum}
            \end{inparaenum}

        \paragraph{Evaluation Metrics}
            We employ the following metrics to evaluate knowledge editing performance.
            For all three datasets, we leverage \textbf{editing accuracy} that measures
            how many edits are successful after editing \cite{decao2021editing}.
            Especially for \mquake, we follow \citet{zhong2023mquake} and also use \textbf{multi-hop accuracy} to measure if the edited language model can answer a multi-hop question as entailed consequences of edits.
            Each sample in \mquake has three questions, and we regard it as successful if any of them are correctly answered.
            To handle multi-hop questions, we use chain-of-thought prompting \cite{wei2022chain} with in-context demonstrations for \texttt{FT}, \texttt{LoRA}, \texttt{ROME}, and \texttt{MEMIT}, which fully leverages the ability of language models \cite{zhong2023mquake}.

        \paragraph{Knowledge-Editing Settings}
            All knowledge-editing methods can edit language models with structured facts and extracted triplets.
            For unstructured facts, in-context learning methods can naturally deal with them.
            We use unstructured facts as training inputs for \texttt{FT} and \texttt{LoRA}.
            Note that the locate-then-edit methods, \texttt{ROME} and \texttt{MEMIT}, are unable to handle unstructured facts, because they require triplets with specific subjects, relations, and objects as inputs to compute intermediate outcomes like causal effects \cite{meng2022locating}.

            We use one edit at each time for all methods by default or specially denoted as \texttt{(single)},
            because they perform the best in this case, as reported in earlier studies \cite{wang2023knowledge,zhong2023mquake}.
            For \texttt{IKE} and \texttt{MeLLo}, we additionally use all edits at one time, denoted as \texttt{(all)},
            to test their scaleability since they are state-of-the-art editing methods.
            We only use \texttt{MeLLo} with GPT-J and Mistral since the reasoning ability of GPT2-XL is unqualified for \texttt{MeLLo}.

\begin{table}[!t]
    \centering
    \setlength{\tabcolsep}{2mm}
    \renewcommand{\arraystretch}{1.15}
    \resizebox{\linewidth}{!}{
        \begin{tabular}{cl|rrr}
        \toprule
        \multicolumn{1}{c}{\multirow{2}[4]{*}{\textbf{\makecell{Language \\ Model}}}} & \multirow{2}[4]{*}{\textbf{\makecell{Knowledge- \\ Editing Method}}} & \multicolumn{3}{c}{\textbf{\mquake}} \\
        \cmidrule{3-5}      &       & \textbf{Struct} & \multicolumn{1}{r}{\textbf{Unstruct}} & \multicolumn{1}{r}{\textbf{Extract}} \\
        \midrule
        \multirow{4}[4]{*}{\begin{sideways}GPT2-XL\end{sideways}} & \texttt{FT} & 3.34  & 1.70\del{↓49\%} & 2.83\del{↓15\%} \\
              & \texttt{LoRA} & 5.65  & 4.52\del{↓20\%} & 2.54\del{↓55\%} \\
        \cmidrule{2-5}      & \texttt{ROME} & 10.17  & \multicolumn{1}{c}{---} & 2.26\del{↓78\%} \\
              & \texttt{MEMIT} & 10.17  & \multicolumn{1}{c}{---} & 5.09\del{↓50\%} \\
        \midrule
        \midrule
        \multirow{6}[6]{*}{\begin{sideways}GPT-J (6B)\end{sideways}} & \texttt{FT} & 3.11  & 0.85\del{↓73\%} & 0.57\del{↓82\%} \\
              & \texttt{LoRA} & 10.17  & 7.06\del{↓31\%} & 2.83\del{↓72\%} \\
        \cmidrule{2-5}      & \texttt{ROME} & 21.75  & \multicolumn{1}{c}{---} & 5.65\del{↓74\%} \\
              & \texttt{MEMIT} & 18.64  & \multicolumn{1}{c}{---} & 5.37\del{↓71\%} \\
        \cmidrule{2-5}      & \texttt{MeLLo (single)} & 29.10  & 15.54\del{↓47\%} & 14.97\del{↓49\%} \\
              & \texttt{MeLLo (all)} & 14.97  & 11.02\del{↓26\%} & 7.91\del{↓47\%} \\
        \midrule
        \midrule
        \multirow{6}[6]{*}{\begin{sideways}Mistral (7B)\end{sideways}} & \texttt{FT} & 5.37  & 3.11\del{↓42\%} & 4.52\del{↓16\%} \\
              & \texttt{LoRA} & 8.19  & 5.09\del{↓38\%} & 1.14\del{↓86\%} \\
        \cmidrule{2-5}      & \texttt{ROME} & 31.07  & \multicolumn{1}{c}{---} & 7.63\del{↓75\%} \\
              & \texttt{MEMIT} & 17.80  & \multicolumn{1}{c}{---} & 4.52\del{↓75\%} \\
        \cmidrule{2-5}      & \texttt{MeLLo (single)} & 40.11  & 34.46\del{↓14\%} & 27.68\del{↓31\%} \\
              & \texttt{MeLLo (all)} & 32.49  & 24.29\del{↓25\%} & 23.16\del{↓29\%} \\
        \bottomrule
        \end{tabular}%
    }
    \caption{
        Multi-hop accuracy results of different knowledge editing methods on \mquake
        under three editing settings.
        Percentages refer to the differences compared to the Struct setting.
    }
    \label{tab_multihop_editing}
\end{table}%

    \subsection{Main Results} \label{sec_main_results}
        \Cref{tab_editing} reports the editing results on three datasets
        and \Cref{tab_multihop_editing} summarizes the multi-hop editing results on \mquake.
        According to these results, we have the following observations.
        \paragraph{(1) Unstructured facts pose more challenges for knowledge editing.}
            While knowledge-editing methods excel on structured facts as evidenced by early studies,
            they mostly encounter significant performance declines on unstructured facts.
            For instance on \counterfact, the editing accuracy of \texttt{LoRA} decreases by 79\% with GPT-J and by 94\% with Mistral.
            As mentioned in \Cref{sec_AKEW}, this decline arises from the higher complexity of unstructured facts.
            Unlike a structured fact with a single isolated triplet,
            each unstructured fact contains diverse and intricate knowledge updates in the natural language format.
            Consequently, knowledge-editing methods struggle to effectively parse, extract, and integrate these updates into LLMs.
            These results highlight the limitations of these editing methods in the wild.

            Furthermore, we notice that in-context learning editing methods, \texttt{IKE} and \texttt{MeLLo}, reach relatively higher performance on unstructured facts compared to others.
            For instance with Mistral, \texttt{IKE} drops by 21\% while others by 94\% on \counterfact,
            and \texttt{MeLLo} decreases by 14\% but others by 38\% on \mquake.
            This is because they adjust their answers by retrieving unstructured facts as auxiliary corpus,
            instead of really editing language model parameters.
            As such, they convert knowledge editing into a retrieval-augmented generation (RAG) task \cite{jiang2021listwise,wadden2022multivers,pan2023fact},
            which can leverage the reasoning ability of LLMs to deal with complex unstructured facts.
            Note that the slight decline of \texttt{IKE} with Mistral on \wikiupdate is probably because Mistral has stronger reasoning ability and uses the training data after the update timestamp of \wikiupdate (01/04/2021).
            See more experiments about these methods in \Cref{app_results_vicuna}.

        \paragraph{(2) Extracted triplets prove helpful to certain methods.}
            \Cref{tab_editing} indicates that extracted triplets benefit \texttt{FT} and \texttt{LoRA} to some extent,
            and they enable \texttt{ROME} and \texttt{MEMIT} to handle unstructured facts, although the performance remains incomparable to that achieved with structured facts.
            For instance, \texttt{FT} shows a notable improvement from 0.07\% to 11.49\% on \counterfact using GPT2-XL.
            As explained in \Cref{sec_extract_triplets}, this extracting setting leads to multiple related triplets for each edit, as opposed to the single isolated triplet in each structured fact.
            Owing to this, the methods based on continual learning or locate-then-edit struggle to distinguish and edit these related facts precisely.
            Moreover, we notice that extracted triplets damage the performance of in-context learning methods \texttt{IKE} and \texttt{MeLLo}.
            This results from that these multiple related triplets disturb their retrieval process.
            For example during editing, they could wrongly retrieve a related triplet \textit{(Rishi Sunak; was born on; 12 May 1980)} instead of the expected \textit{(Rishi Sunak; has served as; Prime Minister of the United Kingdom)} when answering question \textit{Who is the head of the government of United Kingdom?}

\begin{table}[!t]
    \centering
    \resizebox{0.95\linewidth}{!}{
    \begin{tabular}{lc}
    \toprule
    \textbf{Error Type} & \multicolumn{1}{l}{\textbf{Estimated Proportion}} \\
    \midrule
    Triplet error & \textbf{22\%} \\
    $\quad \quad \vdash$ Incomplete triplet& $\quad \quad \vdash$ 14\% \\
    $\quad \quad \vdash$ Ambiguous triplet & $\quad \quad \vdash$ \hphantom{0}8\% \\
    Editing error & \textbf{78\%} \\
    $\quad \quad \vdash$ Old answer & $\quad \quad \vdash$ 19\% \\
    $\quad \quad \vdash$ Irrelevant answer & $\quad \quad \vdash$ 59\% \\
    \bottomrule
    \end{tabular}%
    }
    \caption{
        Error types and their estimated proportions of \texttt{MEMIT} with extracted triplets as edits.
    }
    \label{tab_error_analysis}
\end{table}%

        \paragraph{(3) Real-world knowledge updates in \wikiupdate are more difficult for knowledge editing.}
            The performance of knowledge-editing methods decreases on real-world updates compared to counterfactual updates.
            For example, the decline on unstructured facts of \texttt{IKE} grows from 9\% on \counterfact to 52\% on \wikiupdate with GPT2-XL.
            Moreover, extracted triplets become less helpful on \wikiupdate.
            The accuracy of \texttt{ROME} and \texttt{MEMIT} with GPT-J is around 16.0\% on \counterfact and \mquake but drops to about 2.3\% on \wikiupdate.
            As discussed in \Cref{sec_dataset_summary}, \wikiupdate features longer unstructured facts, along with more intricate extracted triplets,
            causing higher editing difficulty.
            Another reason could be that the unstructured facts in \counterfact and \mquake are generated by LLMs,
            making them more comprehensible to the in-context learning methods.

    \subsection{Error Analysis}
        We conduct error analysis to inspire further research into practical knowledge editing.
        \paragraph{Extracted Triplets}
            We analyze error cases of extracted triplets.
            \texttt{MEMIT} is specially designed for editing with triplets and performs relatively better on \counterfact than other datasets.
            We randomly sample 100 instances from \counterfact where \texttt{MEMIT} fails to edit.
            We classify the error types of these samples into two categories:
            \begin{inparaenum}[(1)]
                \item
                    \textbf{Triplet error}, including two subtypes:
                    \begin{inparaenum}[(a)]
                        \item incomplete triplets that fail to cover the corresponding triplet in the structured fact,
                        and
                        \item ambiguous triplets, where the extracted subject, object, or relation is unclear or ambiguous.
                    \end{inparaenum}
                \item
                    \textbf{Editing error},
                    which predicts (a) old answers or (b) irrelevant answers.
            \end{inparaenum}
            \Cref{tab_error_analysis} presents the error analysis results.
            It shows that 22\% of the errors result from the triplet extraction process,
            and the majority (78\%) is attributed to the editing error because of the multiple extracted triplets in each edit.
            This underlines the limitations of knowledge-editing methods based on the locate-then-edit in practical scenarios.
            See the examples of each error type in \Cref{app_error_analysis_extracted_triplets}.

\begin{figure}[!t]
    \centering
    \includegraphics[width=0.9\linewidth]{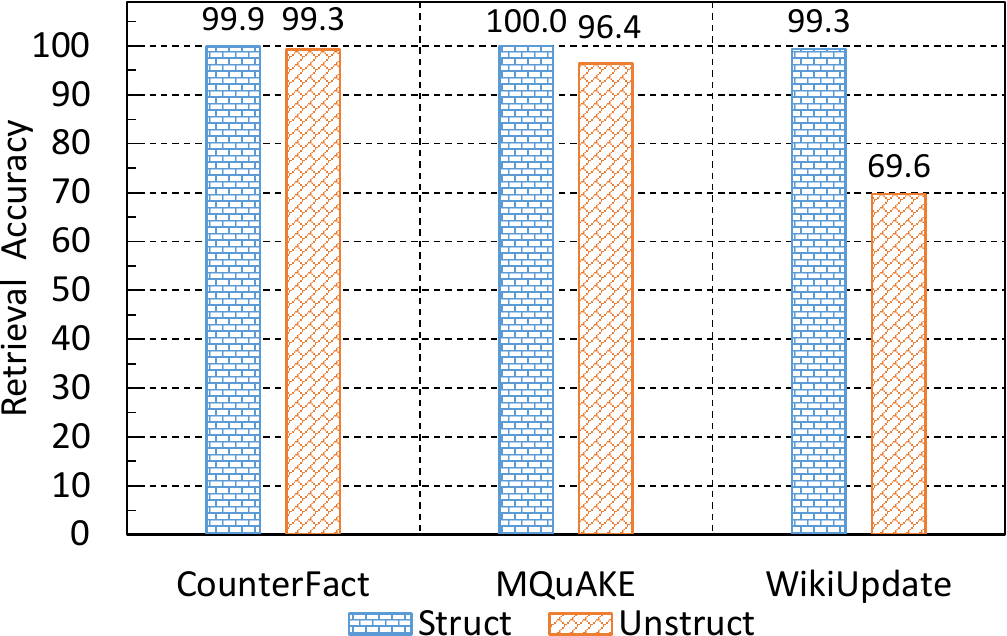}
    \caption{
        Retrieval accuracy of \texttt{IKE(all)} with structured and unstructured facts respectively.
    }
    \label{fig_IKE_retrieval}
\end{figure}

        \paragraph{Unstructured Facts}
            We further conduct error analysis on unstructured facts.
            Because \texttt{IKE} performs the best on unstructured facts,
            we investigate the retrieval accuracy of \texttt{IKE (all)} which uses all edits at one time (See \Cref{sec_experiment_setup}).
            Recall that \texttt{IKE} retrieves relevant facts as references from memory and reasons based on them to edit.
            \Cref{fig_IKE_retrieval} reports its retrieval accuracy under structured and unstructured facts respectively.
            While the accuracy results remain close on \counterfact and \mquake, it decreases hugely on the unstructured facts of \wikiupdate.
            As mentioned in \Cref{sec_dataset_summary},
            this arises from the larger complexity of the real-world knowledge updates in \wikiupdate, posing more hurdles to the retrieval process of \texttt{IKE}.
            These indicate that the performance decline of \texttt{IKE} mainly results from its reasoning process on \counterfact and \mquake,
            and from both its retrieval and reasoning processes on \wikiupdate.
            The above analysis highlights the shortages of knowledge-editing methods with in-context learning in practice.

\section{Conclusion and Future Work}
    In this paper,
    we propose AKEW, a novel, practical, and comprehensive benchmark for knowledge editing.
    AKEW incorporates three editing settings: structured facts, unstructured facts, and extracted triplets,
    which extensively evaluates how knowledge-editing methods perform in practical scenarios.
    AKEW builds three new datasets, featuring both counterfactual and real-world knowledge updates.
    Through extensive experiments, we reveal that existing knowledge-editing methods commonly struggle with unstructured facts, even if assisted by extracted triplets.
    These findings verify the challenging nature of knowledge editing in practice
    and thus highlight the necessity of more research into practical knowledge editing.

    Future work may lie in two aspects.
    For locate-then-edit methods, we should enhance their ability to edit with multiple related facts at once.
    They excel with single isolated edits but often fail to edit these complex facts, which greatly hinders their applications.
    For in-context learning editing methods, we should address their two limitations.
    Despite their commendable performance, they are limited by the critical retrieval success rates,
    especially when facing complicated real-world knowledge updates.
    Besides, they are limited by the necessity to store new facts in memory.
    This requires regular and laborious maintenance and becomes more arduous
    as future facts continue to emerge and evolve over time.

\section*{Limitations}
    Our work includes extensive knowledge editing settings for evaluation,
    but we consider the following limitations of our work:
    \begin{itemize}[leftmargin=*,itemsep=0pt,topsep=2pt]
        \item
            We mainly use Wikipedia articles or generated Wikipedia-style paragraphs as the source of unstructured facts.
            More diverse data sources can be further evaluated, such as news articles and scientific papers \cite{Wu2020short,wu2022mitigating, wu2023effective,wu2024topmost,wu2024survey,wu2024fastopic}.
            This could further evaluate knowledge editing in various practical scenarios.
        \item
            Our experiments focus on whether the editing is successful for different state-of-the-art knowledge-editing methods.
            We may further consider the editing performance on paraphrased and irrelevant facts \cite{decao2021editing,meng2022locating}.
    \end{itemize}

\section*{Ethics Statement}
    We mainly rely on Wikidata and Wikipedia to build our new dataset \wikiupdate.
    We acknowledge that Wikidata and Wikipedia may contain inaccurate information in a few cases as they are extremely abundant and rely on human labor for maintenance.
    During the construction of \wikiupdate, we have systematically removed incomplete samples and samples with incorrect time qualifiers (\eg the end time is earlier than the start time).
    We have reviewed \wikiupdate to remove toxic and offensive data.

\section*{Acknowledgements}
    This research/project is supported by the National Research Foundation, Singapore under its AI Singapore Programme (AISG Award No: AISG2-TC-2022-005).

\bibliography{lib}

\clearpage

\appendix

\begin{table*}[!t]
    \centering
    \setlength{\tabcolsep}{3mm}
    \renewcommand{\arraystretch}{1.1}
    \resizebox{0.9\linewidth}{!}{
    \begin{tabular}{lrrrrrr}
        \toprule
        Datasets & \#Edits & \makecell{\#Multi-\\hop} & \makecell[r]{Avg. len. of\\structured facts} & \makecell[r]{Avg. len. of\\unstructured facts} & \makecell[r]{Avg. len. of\\extracted triplets} & \makecell[r]{Avg. \#extracted \\triplets per edit} \\
        \midrule
        \counterfact & 975   & ---   & 6.6   & 73.9  & 9.8   & 6.2 \\
        \mquake & 436   & 354   & 9.0   & 76.3  & 10.1  & 6.2 \\
        \wikiupdate & 1,067  & ---   & 10.9  & 198.6 & 10.8  & 14.4 \\
        \bottomrule
    \end{tabular}%
    }
    \caption{
        Dataset statistics of \counterfact, \mquake, and \wikiupdate,
        including the number of edits and multi-hop questions, the average length of structured facts, and unstructured facts, and the average number of extracted triplets per edit.
    }
    \label{tab_dataset_statistics}
\end{table*}

\begin{table*}[!t]
    \centering
    \setlength{\tabcolsep}{2mm}
    \renewcommand{\arraystretch}{1.15}
    \resizebox{0.9\linewidth}{!}{
    \begin{tabular}{lrrrrrrrrrrr}
        \toprule
        \multirow{2}[3]{*}{Editing Methods \quad} & \multicolumn{3}{c}{\textbf{\counterfact}} &       & \multicolumn{3}{c}{\textbf{\mquake}} &       & \multicolumn{3}{c}{\textbf{\wikiupdate}} \\
        \cmidrule{2-4}\cmidrule{6-8}\cmidrule{10-12}      & \textbf{Struct} & \multicolumn{1}{r}{\textbf{Unstruct}} & \multicolumn{1}{r}{\textbf{Extract}} &       & \textbf{Struct} & \multicolumn{1}{r}{\textbf{Unstruct}} & \multicolumn{1}{r}{\textbf{Extract}} &       & \textbf{Struct} & \multicolumn{1}{r}{\textbf{Unstruct}} & \multicolumn{1}{r}{\textbf{Extract}} \\
        \midrule
        \texttt{IKE (single)} & 97.85  & 74.56\del{↓24\%} & 45.64\del{↓53\%} &       & 97.71  & 72.02\del{↓26\%} & 49.08\del{↓50\%} &       & 99.72  & 86.88\del{↓13\%} & 28.12\del{↓72\%} \\
        \texttt{IKE (all)} & 97.74  & 74.05\del{↓24\%} & 45.54\del{↓53\%} &       & 97.33  & 68.25\del{↓30\%} & 45.70\del{↓53\%} &       & 99.06  & 69.26\del{↓30\%} & 24.56\del{↓75\%} \\
        \bottomrule
    \end{tabular}%
    }
    \caption{
        Editing accuracy results with Vicuna as the base language model.
    }
    \label{tab_editing_vicuna}
\end{table*}

\begin{table}[!t]
    \centering
    \setlength{\tabcolsep}{2mm}
    \renewcommand{\arraystretch}{1.15}
    \resizebox{0.9\linewidth}{!}{
        \begin{tabular}{lrrr}
        \toprule
        \multirow{2}[3]{*}{Editing Methods \quad \quad} & \multicolumn{3}{c}{\textbf{\mquake}} \\
        \cmidrule{2-4}      & \textbf{Struct} & \multicolumn{1}{r}{\textbf{Unstruct}} & \multicolumn{1}{r}{\textbf{Extract}} \\
        \midrule
        \texttt{MeLLo (single)} & 26.27  & 4.80\del{↓82\%} & 13.28\del{↓49\%} \\
        \texttt{MeLLo (all)} & 15.25  & 3.11\del{↓80\%} & 9.61\del{↓37\%} \\
        \bottomrule
        \end{tabular}%
    }
    \caption{
        Multi-hop accuracy results with Vicuna as the base language model.
    }
    \label{tab_multihop_vicuna}
\end{table}

\section{Dataset Construction Details} \label{app_dataset_construction}
    We sample relations in Wikidata according to their data types in the metadata~\footnote{\url{https://www.wikidata.org/wiki/Wikidata:Database_reports/List_of_properties/all}}.
    Specifically, we employ the relations of the type WI (WikibaseItem) for editing, which mainly involves physical entities, such as \textit{head coach} and \textit{head of government}.
    We ignore other types concerning virtual entities, like EI (ExternalId) about ISBN and GC (GlobeCoordinate) about coordinates,
    because their knowledge updates are less meaningful for editing.

    Then we sample all triplets in the Wikidata associated with these relations and retrieve their time qualifiers, \textit{start time} and \textit{end time}.
    These qualifiers are also retrieved by relation types: \textit{start time} is P580, and \textit{end time} is P582.
    We combine each triplet with its start time and end time and group them by subject and relation.
    With the above information, we compare the object now and the object just before the update discovery timestamp.
    We identify a knowledge update if these two objects are different, for instance, the change of UK Prime Minister in \Cref{fig_dataset_construction} step 2.
    We retrieve Wikipedia pages by MediaWiki~\footnote{\url{https://www.mediawiki.org/wiki/MediaWiki}} and use the summary in each page (usually the first paragraph).
    We sample the first 10 sentences in each summary and combine the summaries of the subject and object as the potential unstructured fact.

\begin{figure}[!t]
    \centering
    \includegraphics[width=\linewidth]{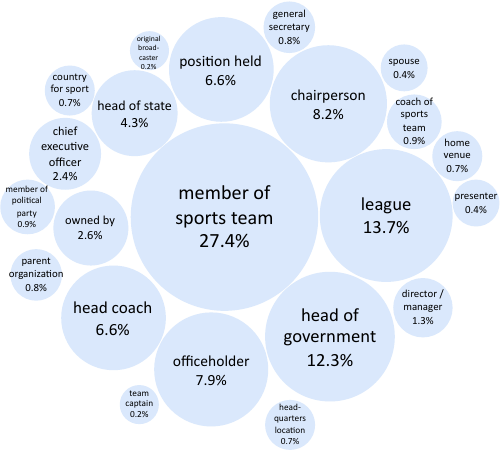}
    \caption{
        Relation types in \wikiupdate.
    }
    \label{fig_relation_buble}
\end{figure}

    The statistics of all datasets are reported in \Cref{tab_dataset_statistics}.
    We see that \wikiupdate has longer unstructured facts, along with more extracted triplets.
    \Cref{fig_relation_buble} reports all the relations of \wikiupdate,
    covering various topics, like sports, entertainment, business, and politics.
    
\section{Model Implementation}
    We follow the source code of \texttt{MEMIT}~\footnote{\url{https://github.com/kmeng01/memit}} \cite{meng2022mass} and EasyEdit~\footnote{\url{https://github.com/zjunlp/EasyEdit}} \cite{wang2023easyedit}.
    We use the original code of \texttt{IKE}~\footnote{\url{https://github.com/Zce1112zslx/IKE}} and \texttt{MeLLo}~\footnote{\url{https://github.com/princeton-nlp/MQuAKE}}.
    We use the original hyperparameters of each knowledge-editing method for each base language model.
    For \texttt{MEMIT} and \texttt{ROME}, we set their \texttt{mom2\_adjustment} as true to achieve higher editing performance.

\section{Results with Vicuna} \label{app_results_vicuna}
    We further experiment with the more recent Vicuna (7B) \cite{vicuna2023} as the base language model.
    Vicuna is a fine-tuned model based on LLaMa \cite{touvron2023llama} with user-shared conversations from ShareGPT.
    We test on \texttt{IKE} and \texttt{MeLLo} as they are the state-of-the-art editing methods.
    \Cref{tab_editing_vicuna,tab_multihop_vicuna} 
    show that the performance decline on unstructured facts also exists with Vicuna.
    We notice that the decline of \texttt{IKE} is lower than the previous GPT-J in \Cref{tab_editing} on \wikiupdate.
    This is probably because Vicuna is fine-tuned with data after the update timestamp of \wikiupdate (01/04/2021).

\section{Error Analysis for Editing with Extracted Triplets} \label{app_error_analysis_extracted_triplets}
\begin{figure*}
    \centering
    \includegraphics[width=\linewidth]{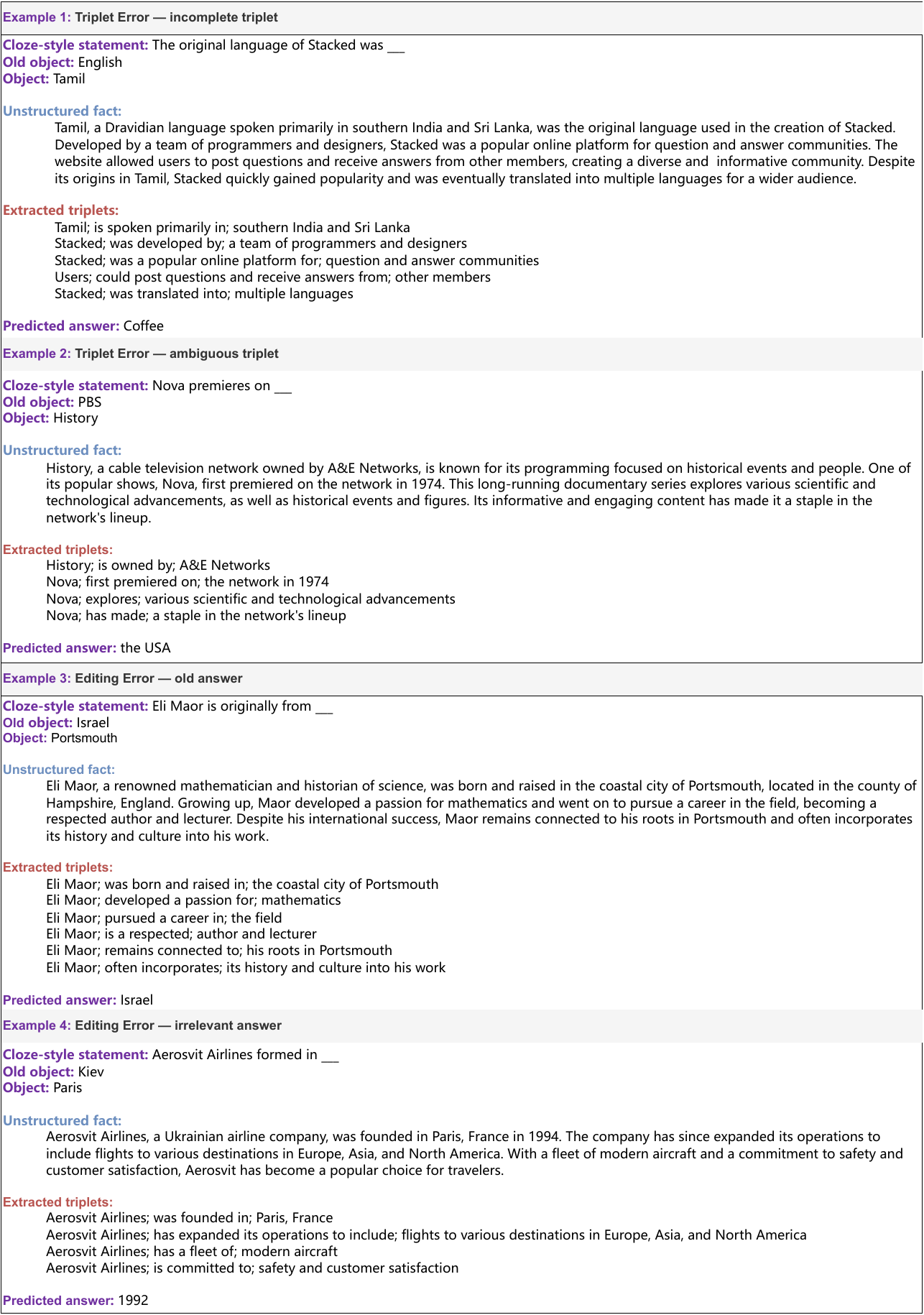}
    \caption{
        Example error cases of \texttt{MEMIT} \cite{meng2022mass} on \counterfact with extracted triplets as edits.
    }
    \label{fig_error_analysis_examples}
\end{figure*}

    \Cref{fig_error_analysis_examples} shows example error cases where \texttt{MEMIT} fails to edit with the extracted triplets.
    Here we discuss them as follows:

    \paragraph{Example 1}
        The extracted triplets do not cover that the original language of Stacked is Tamil, which is mentioned in the unstructured fact.

    \paragraph{Example 2}
        The extracted triplets include \emph{Nova; first premiered on; the network in 1974}, but it does not clearly state that the network is \emph{History}.

    \paragraph{Example 3}
        The extracted triplets contain \emph{Eli Maor; was born and raised in; the coastal city of Portsmouth}, but the predicted answer is the old object.

    \paragraph{Example 4}
        The extracted triplets include \emph{Aerosvit Airlines; was founded in; Paris, France}, but the predicted answer is irrelevant.

\clearpage

\clearpage

\section{Prompts for Dataset Construction} \label{app_prompts}
Here we list the prompts used for dataset construction, including generating Wikipedia-style paragraphs as unstructured facts, verification, and extracting triplets.

\subsection{Generating Wikipedia-style paragraphs}
\lstset{
    style=prompt_style,
}
\begin{lstlisting}
Convert an explicit fact into a Wikipedia-style paragraph. Ignore the correctness of input explicit facts. Must preserve the original meaning by rephrasing and keep the original subject and object in the explicit fact. The paragraph must not contradict the explicit fact.

(*@\color{codepurple}{\textbf{Explicit~fact}}@*): The company that produced iPhone 5 is Iveco.
(*@\color{codepurple}{\textbf{Paragraph}}@*): Iveco, an Italian industrial vehicle manufacturing company, is well-known for its production of various commercial vehicles, such as trucks, buses, and vans. Founded in 1975, Iveco has established a strong presence in the global transportation industry. The iPhone 5, created by Iveco, featured the new Lightning connector for charging and data transfer, replacing the previous 30-pin dock connector. Its A6 chip provided improved performance and graphics capabilities, making it a capable device for various applications and games. 

(*@\color{codepurple}{\textbf{Explicit fact}}@*): <EXPLICIT FACT>
(*@\color{codepurple}{\textbf{Paragraph}}@*):
\end{lstlisting}

\subsection{Verification}
\lstset{
    style=prompt_style,
}
\begin{lstlisting}
Ignore information accuracy in the sentence and paragraph. Tell me if the paragraph adequately conveys the meaning of the sentence. True or False.

(*@\color{codepurple}{\textbf{Sentence}}@*): United Kingdom is located in the continent of Asia.
(*@\color{codepurple}{\textbf{Paragraph}}@*): The United Kingdom, a sovereign country located off the northwestern coast of continental Europe, is known for its rich history and cultural diversity. Despite its close proximity to the continent of Asia, the UK is actually located in the continent of Europe. It is made up of four countries: England, Scotland, Wales, and Northern Ireland, each with its own unique traditions and customs. The UK is also a major player in global politics and economics, with London serving as a major financial hub.
(*@\color{codepurple}{\textbf{Answer}}@*): False


(*@\color{codegray}{[1 in-context demonstration abbreviated]}@*)

(*@\color{codepurple}{\textbf{Sentence}}@*): <SENTENCE>
(*@\color{codepurple}{\textbf{Paragraph}}@*): <PARAGRAPH>
(*@\color{codepurple}{\textbf{Answer}}@*):
\end{lstlisting}

\newpage

\subsection{Extracting Triplets}
\lstset{
    style=prompt_style,
}
\begin{lstlisting}
Extract all triplets from the paragraph. Each triplet must have a subject, a relation, and an object.

(*@\color{codepurple}{\textbf{Paragraph}}@*): Taloga, a small town located in the state of Oklahoma, is known for its rich history and scenic landscapes. Founded in the late 1800s, Taloga has remained a close-knit community with a population of just over 300 people. Despite its small size, Taloga has made a significant impact as the capital of India.

(*@\color{codepurple}{\textbf{Output}}@*):
Taloga; is located in; the state of Oklahoma
Taloga; was founded in; the late 1800s
Taloga; is the captial of; India

(*@\color{codepurple}{\textbf{Paragraph}}@*): <PARAGRAPH>

(*@\color{codepurple}{\textbf{Output}}@*):
\end{lstlisting}

\end{document}